\documentclass[sigconf,authorversion,nonacm]{acmart}

\author{Yongye Su}
\email{su311@purdue.edu}
\affiliation{%
  \institution{Department of Computer Science, Purdue University}
  \city{West Lafayette}
  \state{Indiana}
  \country{USA}
}

\author{Yuqing Wu}
\email{yuqingw1@uchicago.edu}
\affiliation{%
  \institution{Applied Data Science, University of Chicago}
  \city{Chicago}
  \state{Illinois}
  \country{USA}
}
\usepackage{blindtext}

\usepackage{etoolbox}
\begin{document}

\title{Robust Detection of LLM-Generated Text: A Comparative Analysis}



\begin{abstract}
The ability of large language models to generate complex texts allows them to be widely integrated into many aspects of life, and their output can quickly fill all network resources. As the impact of LLMs grows, it becomes increasingly important to develop powerful detectors for the generated text. This detector is essential to prevent the potential misuse of these technologies and to protect areas such as social media from the negative effects of false content generated by LLMS. The main goal of LLM-generated text detection is to determine whether text is generated by an LLM, which is a basic binary classification task. In our work, we mainly use three different classification methods based on open source datasets: traditional machine learning techniques such as logistic regression, k-means clustering, Gaussian Naive Bayes, support vector machines, and methods based on converters such as BERT, and finally algorithms that use LLMs to detect LLM-generated text. We focus on model generalization, potential adversarial attacks, and accuracy of model evaluation. Finally, the possible research direction in the future is proposed, and the current experimental results are summarized.
\end{abstract}

\maketitle


\section{Introduction}
The advent of Large Language models (LLMS) marks a new evolution in text generation technology, making it more pervasive and human-like than ever before. The LLM model is built on the basis of deep learning architecture, mainly using Transformer model, and is trained on a large scale text corpus to provide complex language sentence patterns and emotional details. These models leverage self-attention mechanisms to emphasize different pieces of text when predicting subsequent words, enabling nuance and context-aware text generation. In their technique, autoregressive models stand out by predicting future markers based on previous contexts, thus maintaining consistency and relevance. In addition, the llm has been fine-tuned in specific areas to improve accuracy and applicability. 

An important development in LLM is the introduction of search-enhanced generation, which uses an external database to inject the latest information into the generated content in real time. This capability not only enriches the knowledge base of the LLM, but also ensures the relevance of the output and the notification of the latest data. However, the rise of LLM has also brought its own set of challenges, particularly in distinguishing between human-generated and machine-generated text. Traditional detection methods rely on simple statistical patterns and are difficult to match the complex output of llm, which closely mimics human writing style and structure.

This situation necessitates the development of advanced detection mechanisms that can comprehend and dissect the deep contextual embeddings and intricacies unique to LLM-generated texts. These new methodologies need to evolve in tandem with LLM advancements, incorporating insights into token prediction patterns, stylistic tendencies, and the integration of retrieval-based content. As we forge ahead, it is crucial to develop strategies that ensure the authenticity and integrity of digital content in the age of advanced language models.

This situation requires the development of advanced detection mechanisms capable of understanding and dissecting the deep contextual embeddedness and complexity characteristic of LLM-generated text. At the same time, in an ideal state, it can synchronize and iterate with the development of llm, and play a high effect in maintaining network security, promoting information authenticity and improving user experience. Recognize and process all kinds of text content, such as spam, phishing, and other malicious messages, thereby protecting users from cyber attacks and information fraud.

\subsection{Key questions that we have answered}
\begin{enumerate}
    \item Among traditional regression methods, standard neural networks, and approaches like BERT, which is more suitable for this task?
    \item Will our method still be effective if faced with datasets that have different features from the current one?
    \item How do we address data ambiguity? This issue is closely related to the inherent mechanisms of detection technologies. The widespread deployment of LLMs across various fields has made it increasingly difficult to discern whether training data was written by humans or generated by LLMs.
\end{enumerate}
\section{Background}

With the great developments in artificial intelligence (AI), the significant increase in its various applications benefited human society including economics, medical sciences, and so on.~\cite{su2022yolo} One of the most recent eye-catching developments in this field is generative AI (GenAI) empowered by large language models, such as ChatGPT by OpenAI~\footnote{\url{https://chat.openai.com/}}, Gemini by Google~\footnote{\url{https://gemini.google.com/}}, LLaMA by Meta~\footnote{\url{https://llama.meta.com/}}, etc. Those large language models are well-trained on human language resources like Wikipedia, Reddit, Quora, etc. By heavy training on those human language datasets, the large language models become experts in the style of human language, especially in terms of grammaticality, fluency, coherency, and usage of real-world knowledge without being easily noticed by humans. Moreover, they can generate various kinds of language, thus their text generation is served to the world.

Such a blooming growth in generative AI also attracted more concerns, especially those people with very limited background in AI/ML, who are more likely to be misguided by machine-generated text. Research on the detection of LLM-generated text has received lots of attention even before the advent of ChatGPT~\cite{achiam2023gpt}, especially in areas such as early identification of deepfake text~\cite{10179387} and machine-generated text detection.

While with the groundbreaking advances in generative artificial intelligence (AI) large language models (LLMs) like ChatGPT, people nowadays are heavily using them due to their human-like generated content with intelligence and efficiency, whereas more and more contents that appear online are generated by AI and some of them are non-sense information partly due to the problem called hallucination of LLMs, and partly because of its outdated knowledge base without knowledge from a specific source like vector databases~\cite{jing2024large, su2024vexless}. The misused generated information without any warnings or tags would have been a devastating issue for users without attention or corresponding knowledge, it not only can misguide the problem but also can lead to other potential concerns, like academic ethics, e.g. plagiarism and fabrications~\cite{ji2023survey}.

\subsection{The Rationale for Detection}
Machine-generated text, also tagged as AI-generated text online, has been found in numerous applications, from automated customer service to personalized content creation. However, the increasing indistinguishability of AI-authored text from human writing harbors risks, particularly for those unversed in AI. Machine-generated content can mislead audiences, propelling concerns over the potential for disinformation, especially when it circulates without appropriate disclosure.

While the benefits of LLMs are clear, unchecked expansion of their output could breed adverse outcomes, such as bias reinforcement and negative misinformation~\cite{kreps2022all}. The current version requires content platforms and educational institutions to deploy effective detection tools. These tools must navigate the evolving landscape of LLM capabilities, distinguishing AI-generated content to prevent issues like academic dishonesty and the spread of false narratives across platforms including LinkedIn, Twitter, and Reddit~\cite{sadasivan2023can,mitchell2023detectgpt,liu2024community}. Their potential for misuse by nefarious actors is a critical concern, with instances of fake news propagation, generation of hypocritical product reviews, and spam activities demonstrating the dimmer aspects of this technology. The subtlety with which LLM crafts such deceptive content often renders human detection to mere chance, underscoring the need for advanced models capable of discerning LLM-authored texts from those penned by humans.

Thus, LLMs are double-edged swords in today's digital ecosystem. While they have the capacity to reflect and magnify societal biases—such as gender, racial, or religious prejudices—they also provide a window into understanding these biases through analysis of model outputs. The presence of biases in TGMs is not only a reflection of their training data but can also perpetuate harm and reinforce stereotypes, affecting individuals and groups negatively.

These developments necessitate robust, discerning models for content moderation on vulnerable platforms, including social networks, email services, government portals, and e-commerce sites. Such systems not only protect against misinformation but also uphold integrity and trust in online communications.

\subsection{Large Language Models and Text Generation Methodologies}

The advent of Large Language Models (LLMs) has dramatically altered the story of text generation, rendering it more pervasive and sophisticated. The core methodology underlying LLMs involves training on vast corpora of text data using deep learning architectures—predominantly Transformer models. These architectures facilitate the learning of complex patterns and nuances in language through self-attention mechanisms that weigh the significance of different parts of text when predicting the next word in a sequence.

LLMs employ a variety of techniques to generate text. One fundamental method is the use of autoregressive models that predict subsequent tokens based on the preceding context, allowing for the generation of coherent and contextually relevant content. The more advanced LLMs utilize fine-tuning methods on specific domains or tasks to tailor the generative process, enhancing relevance and accuracy.

Retrieval-augmented generation adds another layer of complexity, where the LLM accesses an external knowledge base to incorporate factual and up-to-date information into the generated text. This not only provides the LLM with a broader knowledge base from which to draw but also allows it to produce content that is context-aware and dynamically informed by the latest available data~\cite{jing2024large, su2024vexless}. Despite the remarkable capabilities of LLMs, their widespread application in generating text has led to challenges in distinguishing machine-generated content from human writing. Conventional detection algorithms, typically based on simpler statistical patterns, face difficulties in classifying outputs from LLMs due to their nuanced understanding and emulation of human language style and content structure.

The emergent need is for enhanced detection mechanisms capable of understanding and analyzing the deep contextual embeddings and intricacies of LLM-generated text. These methodologies must evolve concurrently with LLMs, incorporating understanding of model-specific traits such as token prediction patterns, stylistic tendencies, and the potential inclusion of retrieval-based information. As LLMs continue to evolve, the approaches must be employed to ensure the authenticity and integrity of digital content as well.

\subsection{Zero-Shot Machine-Generated Text Detection}
This study is on the zero-shot detection of machine-generated text—approaching the problem without pre-existing datasets of human or AI-authored samples for training. We are interested in this topic because this method aligns with real-world scenarios where both the versions and variations of LLMs evolve rapidly, and those pre-existing datasets may not be instantly available. Our approach seeks to develop versatile, adaptive models that remain robust across the changing tides of LLM outputs, ensuring reliability and generalizability in a scenario where machine text can be constantly detected, thus such service could be distributed on different general cases.

\section{Related work}
The current state-of-the-art related work for diverse LLM detection includes but is not limited to watermarking technology, for example, secret key-based watermarking technology encompasses the utilization of unique keys (or key pairs) to insert specific signals into generated text or decoded probability vectors. Other interesting methods include fine-tuning LLMs to distinguish the AI-generated content, advising learning methods~\cite{koike2023outfox}, and employing LLMs as detectors~\cite{bhattacharjee2023fighting}. Moreover, it remains a good story for us how traditional regression methods, standard neural networks, and Transformers like BERT suit this task.

First of all, state-of-the-art detectors exhibit considerable performance degradation when faced with out-of-distribution (OOD) data that is general in real-world situations, with the efficacy of some classifiers marginally surpassing that of random classification. An intuitive example is that recent research~\cite{liang2023gpt} revealed a noticeable decline in the performance of state-of-the-art detectors when processing texts authored by non-native English speakers.

On the other hand, with the fast-growing versions of LLMs, the semantic and language styles of LLMs are likely to change over time. We have not yet determined whether current detection models are adequate for changing styles of LLMs. 

Last but not least, some previous research on understanding the capacity of humans to detect AI/machine-generated text has mostly treated the task as a classification task. Given a text example that is either entirely human-written or wholly machine-generated, annotators must predict whether it is human-written or machine-generated, which does not provide much explainability to the task.

\paragraph{\textbf{Supervised Methods:}} Supervised detection methods often rely on labeled datasets to train models that can distinguish between human and LLM-generated text. Techniques have leveraged a variety of machine learning algorithms, from classic decision trees and support vector machines to more complex deep learning structures. Recently, Transformer-based models, such as BERT and its variants, have been fine-tuned for this task, exploiting their deep contextual understanding to identify subtle nuances indicative of AI-generated text. However, their reliance on large, well-labeled datasets limits their applicability in situations where such data are scarce or non-representative of real-world distribution.

\paragraph{\textbf{Unsupervised Methods:}} Unsupervised learning, in contrast, does not require labeled data, and instead, methods such as clustering algorithms and autoencoders are used to discern patterns that may differentiate humans from LLM-generated content. Anomaly detection is one area that has shown promise, where statistical deviations from established norms of human-written text are flagged. These methods benefit from not needing annotated data, yet they may struggle with the high variability and sophistication of text produced by cutting-edge LLMs.

\paragraph{\textbf{Zero-Shot Methods:}} Zero-shot methods stand out for their ability to generalize to unseen examples without direct training on specific examples of the target classification task. In the context of LLM-generated text detection, zero-shot approaches aim to leverage the inherent capabilities of LLMs to recognize their kindred outputs, essentially using LLMs to detect text produced by similar models. This approach's strength lies in its model-irreligious nature, declaring a versatile framework that could adapt to the evolutionary pace of generative AI.

These emerging methods signal a paradigm shift in AI-generated text detection. Still, they each bear intrinsic challenges that must be addressed to achieve reliable and robust systems. The watermarking approach relies on the presence of the watermark, supervised methods require extensive and diverse training data, unsupervised methods need careful tuning to capture the complexities of LLM outputs, and zero-shot methods must effectively harness the transfer learning capabilities of LLMs. Future work must continue to refine these approaches, ensuring they remain effective against the continuously improving quality of text generation models.

\section{Dataset and settings}

In our study, we used the Hello SimpleAI HC3 dataset~\cite{guo-etal-2023-hc3}, and it consists of a different collection of text samples. This dataset is specifically designed to test the benchmark for the performance of various machine learning models which used for differentiating and classifying between human-written text and AI generated answers. The dataset covers a wide range of topics and writing styles. It was selected from informal social media posts like Reddit to formal answers from specialists, therefore it can ensure  we have a comprehensive assessment.

To ensure that we can get complete and appropriate results, we divided the data into three parts. The first part is training subset, another part is validation subsets which can check how well it's running and making adjustments to it, and the last part is a testing subset which used for doing a final check to make sure it's running correctly.

The text samples were preprocessed to normalize factors such as text length, formatting, and encoding, which might otherwise skew the models' performance. Additionally, metadata was stripped to prevent models from using any data-specific identifiers rather than the linguistic features of the text.
\section{Methodologies and Experimental studies}

Our approach used three main machine learning paradigms to tackle the detection of LLM-generated text: Transformer models, traditional machine learning models, and methods specifically for LLM detection.

\subsection{Transformer-based Evaluation}

We employed BERT (Bidirectional Encoder Representations from Transformers) due to its state-of-the-art performance in various NLP tasks. Fine-tuning was carried out on the training subset of the Hello-SimpleAI HC3 dataset, adjusting the model to effectively recognize the subtle differences between human and LLM-generated text.

\subsection{Machine Learning-based Evaluation}

Traditional ML algorithms were applied, including Logistic Regression, Support Vector Machines (SVM), Gaussian Naive Bayes with Bayesian Optimization, and Random Forests. Each model was trained and optimized to capture statistical patterns and inconsistencies in the text that could indicate LLM authorship.

\subsection{LLM-specific Detection Evaluation}

Specialized detection models, such as DetectGPT and Single-revise, were evaluated for their effectiveness in identifying LLM-originated content. These models leverage insights specific to LLM output characteristics and are thus hypothesized to provide a more targeted approach to detection. This indicates a method that comes from the rationale: \textit{While LLMs are generating texts, why can’t we give them back their own way?}

\subsection{Experimental Setup and Evaluation}\label{sec:evaluation}

Experiments were conducted in a controlled environment. Each model underwent a series of tests, first with the training subset to fine-tune the models and then with the validation subset to adjust for the best hyperparameters. The final evaluation was conducted using the test subset, providing a comprehensive measure of each model's detection capabilities.

The performance of each model was assessed using metrics such as accuracy, precision, recall, and F1 score. Additionally, we evaluated the robustness of each model by introducing adversarial examples into the test data, designed to simulate evasion techniques that could be employed by sophisticated LLMs.

Our experimental studies aimed to determine the relative strengths and weaknesses of each methodology, providing insights into the most effective strategies for detecting LLM-generated text in varying contexts.
\section{Corpus and Data Analysis}

Prior to delving into the experimental evaluation, an extensive corpus analysis was performed on the Hello-SimpleAI HC3 dataset. This analysis provided key insights into linguistic patterns characteristic of human versus LLM-generated text, forming a critical foundation for our methodological approach. The following subsections detail the specific findings from our analysis, each supported by visual data representations.

\subsection{Answer Length Distribution}

First, comparing the answers generated by ChatGPT with the answers written by humans. It was found that there was a significant difference in the length of the answers generated by ChatGPT. Figure 1 shows the distribution indicate where ChatGPT's answers tend to be longer. 

\begin{figure}[htbp]
    \centering
    \includegraphics[width=0.5\textwidth]{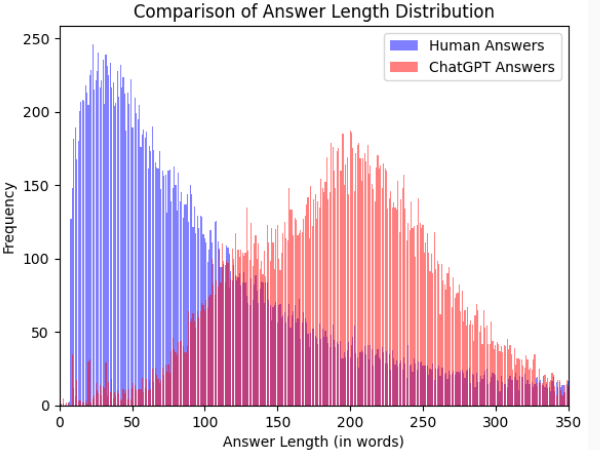}
    \caption{Comparison of Answer Length Distribution}
    \label{fig:llm4llm}
\end{figure}

\subsection{Sentence Length Analysis}

We also compared the sentence length and it showed that ChatGPT not only produced longer answers than human, but also produced longer sentences on average. Figure 2 shows the distribution of sentence lengths, emphasizing the tendency of extended structures in LLM-generated text. 

\begin{figure}[htbp]
    \centering
    \includegraphics[width=0.5\textwidth]{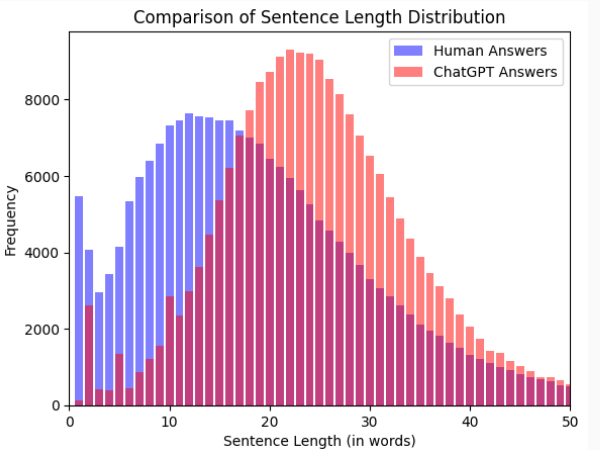}
    \caption{Comparison of Sentence Length Distribution}
    \label{fig:llm4llm}
\end{figure}
\subsection{Vocabulary Diversity}

Then we also examine the diversity of vocabulary used by ChatGPT and humans. It is obvious that humans prefer  a wider range of vocabulary usage, as shown in Figure 3, which plots the frequency of type-to-tag ratio (TTR) ranges.

\begin{figure}[htbp]
    \centering
    \includegraphics[width=0.5\textwidth]{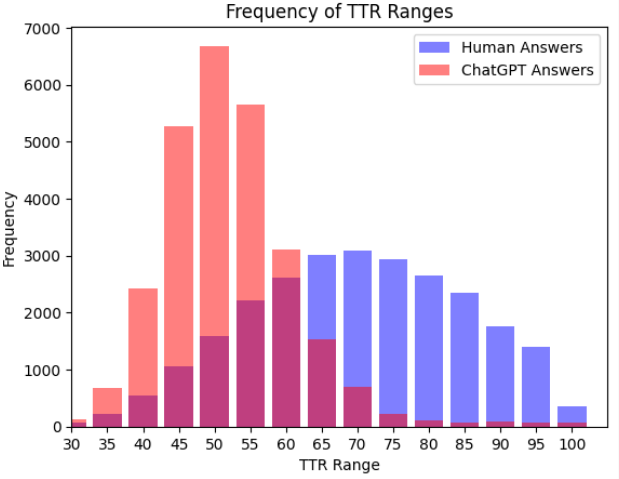}
    \caption{Frequency of TTR Ranges}
    \label{fig:llm4llm}
\end{figure}
\subsection{Lexical Complexity}

The complexity of the words used in the text is another aspect of our analysis. ChatGPT's tend to use more complex vocabulary is evident in Figure 4, which illustrates the distribution of Flesch-Kincaid grade levels.

\begin{figure}[htbp]
    \centering
    \includegraphics[width=0.5\textwidth]{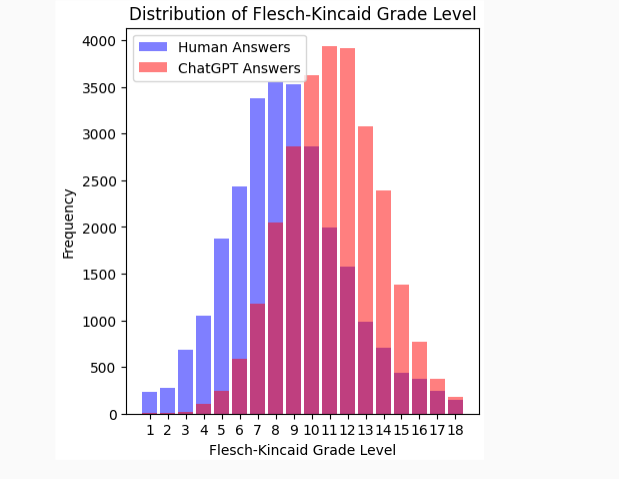}
    \caption{Distribution of Flesch-Kincaid Grade Level}
    \label{fig:llm4llm}
\end{figure}
\subsection{Sentence Structure Complexity}

Finally, the complexity of the sentence structure is assessed by examining the dependency distance in the sentence. The longer dependency distances indicates the more complex the sentence structure it used. As shown in Figure 5, ChatGPT’s sentences exhibit more complex grammatical structures, which indicates that the text generated by LLM has higher complexity

\begin{figure}[htbp]
    \centering
    \includegraphics[width=0.5\textwidth]{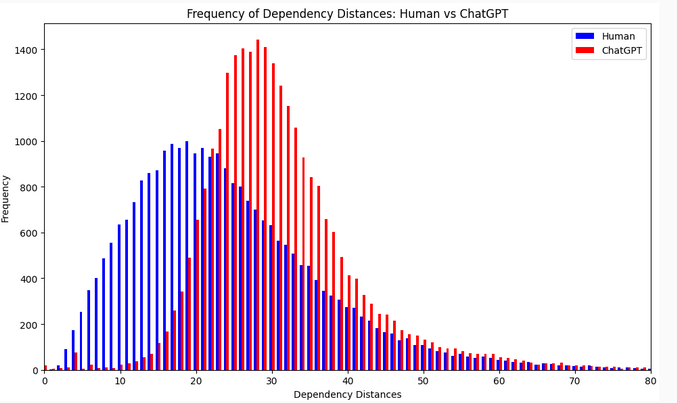}
    \caption{Frequency of Dependency Distances}
    \label{fig:llm4llm}
\end{figure}

These analyzes not only illustrate the different linguistic features of human and LLM generated texts, but also provide key parameters for improving our detection methods.

\section{LLM-generated text Detection Method }
\subsection{Detection Method Based on Machine-learning}

In this project, we primarily use four machine learning techniques to detect text generated by Large Language Models (LLMs): Random Forest, Logistic Regression, Gaussian Naive Bayes, and SVM. Before training with the corresponding models, we preprocessed the dataset using the Word2Vec model for vectorization. Word2Vec, a widely used word embedding technique, was introduced by Google's research team in 2013. Its primary goal is to transform words into vectors in a high-dimensional space, where semantically similar words are closer in vector space. This representation can reveal complex semantic relationships between words, such as synonyms and antonyms.

Word2Vec training typically relies on large-scale text datasets and adjusts word vectors through backpropagation to better capture semantic relationships between words. This method enhances the performance of various natural language processing tasks. 
\subsubsection{Logistic Regression based method }\mbox{}\\
Logistic regression is favored for its simplicity and high interpretability, which is particularly important when a clear explanation of model results is needed. Moreover, logistic regression is efficient during training, allowing for rapid model development even when resources are limited. Most importantly, logistic regression is particularly well-suited for handling high-dimensional sparse data, which is crucial in the detection of text generated by LLMs. In terms of performance evaluation, logistic regression shows moderate accuracy (0.86), recall (0.84), and F1 score (0.85), indicating that it performs well in balancing outcomes, although precision is somewhat lacking. This is because logistic regression assumes that the data is linearly separable, meaning it uses a linear decision boundary to differentiate between categories. If the relationship between features and categories in the dataset is nonlinear, logistic regression may not capture these complex relationships well, impacting model accuracy. Furthermore, logistic regression may perform poorly on high-dimensional data (i.e., when there are many features), especially when there are many irrelevant features or when features exhibit multicollinearity. Although regularization techniques can mitigate this issue, it remains a challenge. Solutions to these drawbacks include using more complex models such as Support Vector Machines (SVM) or Random Forests to handle nonlinear relationships.

\subsubsection{Gaussian Naive Bayes method}\mbox{}\\
The Gaussian Naive Bayes (GNB) algorithm is widely used in text classification tasks due to its good performance in handling large feature spaces, as is often encountered in large language model (LLM) text classifications. This algorithm is particularly suited for scenarios where features are extracted assuming a Gaussian distribution. Gaussian Naive Bayes classifies texts based on probabilistic principles, and this probability-based logic shows advantages in dealing with complex natural language data due to its simplicity.

However, one issue with its application to text data is that the inherent assumption of feature independence in Naive Bayes often does not hold in reality. Despite this, due to the model's simplicity and probabilistic foundation, it still effectively processes and classifies a large volume of text data. Additionally, Gaussian Naive Bayes can experience significant estimation biases when dealing with skewed data distributions.

In terms of performance metrics, Gaussian Naive Bayes has demonstrated high accuracy (0.96) in specific implementations of LLM text classification, but its recall rate (0.81) is slightly lacking, with an F1 score of 0.87. This indicates that although the model is very precise in identifying relevant texts, it falls short in terms of balance.

\begin{table}[h]
\centering
\begin{tabular}{|c|c|c|c|}
\hline
\textbf{Model} & \textbf{Precision} & \textbf{Recall} & \textbf{F1-Score} \\ \hline
SVM            & 0.97               & 0.97            & 0.97              \\ \hline
Logistic Regression & 0.86        & 0.84            & 0.85              \\ \hline
Gaussian Naive Bayes & 0.96        & 0.81            & 0.87              \\ \hline
\end{tabular}
\caption{Model performance comparison}
\end{table}

\subsubsection{SVM based method}\mbox{}\\
In the field of machine learning, SVM is favored for its effectiveness in handling high-dimensional data (e.g., once text data is vectorized). Its ability to combat overfitting is particularly crucial in high-dimensional spaces, often preventing degradation in the model's ability to generalize to new data.

The choice of kernel in SVM significantly affects classification accuracy. The kernel trick allows the model to fit the maximum-margin hyperplane in a transformed feature space. The choice of kernel—linear, polynomial, radial basis function (RBF), or sigmoid—determines the flexibility of the decision boundary.

Training SVM models on large datasets can be computationally expensive, especially with nonlinear kernels. However, Logistic Regression still performs well when the feature space is large, as is common in text classification, partly because it handles sparse data effectively.

For SVM models classifying text generated by LLMs, the precision is 0.97, the recall is 0.97, and the F1 score is 0.97. These metrics indicate that the model is highly precise, with moderate recall, and a balanced overall F1 score, reflecting SVM's effectiveness in text classification scenarios.

\subsection{Detecion Method Based on Transformers}
In this section, we discuss the performance of the transformer-based model BERT in our specific task. The emergence of transformer models has heralded a profound paradigm shift in Natural Language Processing (NLP). Before delving into the specifics of BERT and its applications, it's imperative to understand why transformers have become the backbone of modern NLP tasks and why we perceive its utility in our classification endeavor.

\subsubsection{Advantages of Transformer Models} \mbox{}\\
We typically choose this architecture based on four advantages~\cite{NIPS2017_3f5ee243}:
\begin{itemize}
    \item \textbf{Parallelization}: Unlike RNNs, transformers can process all words in a sequence simultaneously, leading to faster training and inference times.
    \item \textbf{Long-range Dependencies}: Transformers capture dependencies between words irrespective of their distance in the input sequence, which is crucial for understanding context in natural language.
    \item \textbf{Attention Mechanism}: The self-attention mechanism allows transformers to focus on relevant parts of the input sequence, enabling them to capture intricate linguistic patterns effectively.
    \item \textbf{Transfer Learning}: Transformer-based models can be fine-tuned on specific tasks using transfer learning, leveraging pre-trained representations to achieve state-of-the-art performance with minimal data.
\end{itemize}

\subsubsection{Introduction to BERT}\mbox{}\\
Bidirectional Encoder Representations from Transformers(BERT) is a pre-trained transformer-based language model that has revolutionized NLP tasks~\cite{BERT}. Unlike traditional models that process text sequentially, BERT employs a bidirectional approach, where each word is contextualized based on both its preceding and succeeding words. This bidirectional understanding enables BERT to capture deeper semantic meaning and contextual nuances within text sequences. BERT is pre-trained using two unsupervised tasks: Masked Language Model (MLM) and Next Sentence Prediction (NSP). In MLM, random words are masked in the input, and the model is trained to predict the masked words based on the context. In NSP, BERT learns to predict whether two input sentences appear consecutively in a document.

\subsubsection{Model and Fine-Tuning}\mbox{}\\
We opted to use DistilBERT, a distilled version of BERT that offers a smaller memory footprint and faster inference time while retaining most of BERT's performance. For fine-tuning, we freeze the parameters and add a softmax layer as the final classification layer in the network structure and train on the dataset. The fine-tuned DistilBERT on the dataset yielded promising results, with an impressive accuracy of around 99\%. We also applied cross validation and adjusted different train-test ratio and the testing results are all around 97\% accurate. However, upon closer inspection, we observed that the model encountered difficulty in accurately classifying data points outside the confines of our dataset.

\subsubsection{Challenges and Insights}\mbox{}\\
We observed that adding special characters, such as slashes and quotation marks with backslashes, increases the likelihood of the text being classified as AI-generated, raises intriguing questions about the underlying patterns learned by the model. It suggests that the model might be relying heavily on superficial features rather than semantic understanding to make predictions. This phenomenon underscores the importance of robust dataset curation and the potential limitations of relying solely on accuracy as a performance metric.

\subsection{Detecion Method Based on Large Language Models}\mbox{}\\

In this subsection, we will discuss the methodology that utilized LLM to detect LLM-generated text. Specifically, this method's prototype comes from DetectGPT~\cite{mitchell2023detectgpt}, a zero-shot machine-generated text Detection using probability curvature. Inspired by this article, a follow-up work also used a similar approach but a much faster way to use LLM as the keystone of LLM-text detection. Given the original passage $x$, DetectGPT first generates minor perturbations of the passage $x_i$ using a generic pre-trained model such as T5~\cite{raffel2020exploring}. Then DetectGPT compares the log probability under p of the original sample $x$ with each perturbed sample $x_i$. If the average log ratio is high, the sample is likely from the source model~\cite{mitchell2023detectgpt}. 

\textbf{Hypothesis:} To detect LLM-generated text given by users, they have such a hypothesis that minor rewrites of model-generated text tend to have lower log probability under the model than the original sample, while minor rewrites of human-written text may have higher or lower log probability than the original sample. Thus, the log probability vs. different passage $x$ is always at its peak for LLM-generated text.

\textbf{Architecture:} Our modified architecture comes from the prototyped idea from DetectGPT~\cite{mitchell2023detectgpt} and improvised pipeline from~\cite{zhu2023beat}, as shown in Figure~\ref{fig:llm4llm}

\begin{figure}
    \centering
    \includegraphics[width=0.5\textwidth]{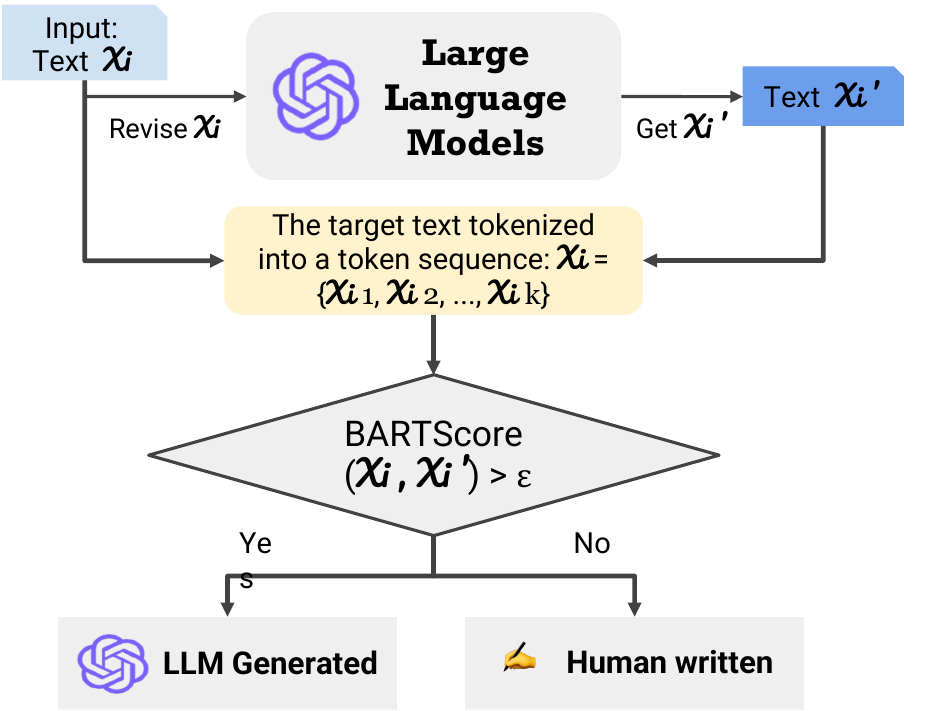}
    \caption{The simplified architecture of using LLM to detect LLM-generated texts.}
    \label{fig:llm4llm}
\end{figure}

Using the evaluation pipeline and methods mentioned in Section~\ref{sec:evaluation} and Figure~\ref{fig:llm4llm}, we successfully conducted experiments with results shown in Table~\ref{tab:llm4llm_comparison}

\begin{table}[b]
    \centering
    \begin{tabular}{|c|c|c|c|}
         Method Name & DetectGPT & Single-Revise \\
         AUROC & 0.952 & 0.943 \\
         Avg. Predict Time & 8.98 & 0.2  
    \end{tabular}
    \caption{The performance and efficiency comparison between the prototype and refined method.}
    \label{tab:llm4llm_comparison}
\end{table}

\section{Extend: Understanding the Illusion of Emergent Abilities in LLMs}
We also delved into the topic of emergent abilities within large language models, specifically through the insights offered by the paper \textit{Are Emergent Abilities of Large Language Models a Mirage}~\cite{NEURIPS2023_adc98a26}.

\subsection{Emergent Abilities}
Recent advancements in AI have brought to light a fascinating phenomenon within large language models (LLMs) known as "emergent abilities." These abilities are seemingly new capabilities that were not observed in smaller-scale models but become apparent as the size and complexity of models increase. The concept of emergent abilities has been explored in various scientific disciplines, suggesting that as systems grow in complexity, new properties can unexpectedly come to the fore. This has been broadly noted in models like GPT-3, PaLM, and LaMDA, causing a debate about the implications of such abilities and their origins.

\subsection{Alternative Perspective}
While emergent abilities have been the subject of much speculation, the paper challenges the conventional understanding of these phenomena. The authors argue that the emergence of such abilities might not be inherent to the AI models themselves. Instead, they might result from the choice of metrics used to evaluate model performance. According to their hypothesis, nonlinear or discontinuous metrics can make model improvements appear sudden and unpredictable, which might be misinterpreted as emergent abilities.

\subsection{The Role of Metrics} The paper posits that when linear or continuous metrics are used, the model's performance improvements appear as smooth, continuous, and predictable, contradicting the notion of emergence. The authors provide a mathematical model to support their claims, showing that certain performance measures can be misleading.

\subsection{Analyzing Model Performance}
 The researchers focus on the GPT family. They use neural scaling laws and demonstrate that the per-token cross-entropy loss decreases in a power-law fashion with the number of model parameters. This finding indicates that the improvements in model performance could be predicted based on the scaling of parameters, challenging the unpredictability aspect of emergent abilities.

 When a nonlinear metric such as Accuracy is employed, performance appears to improve dramatically as the model scales up. However, if the metric is switched to a linear one like Token Edit Distance, this sharp improvement is no longer observed. Instead, the performance enhancement becomes more gradual and predictable. 

 Furthermore, the paper demonstrates that statistical quality affects emergent abilities. By increasing resolution and adding more test data, the authors found that smaller models don’t exhibit zero accuracy but instead experience a predictable decline in performance as target strings grow longer. This suggests that emergent abilities may not inherently result from model scaling.

 \subsection{Implications for Differentiating Human and AI-Generated Text}
 The study concludes that emergent abilities in LLMs may be a construct of evaluation techniques rather than an intrinsic property of AI scaling. This realization is crucial for creating more reliable frameworks for assessing AI performance and differentiating AI-generated text from human writing. Recognizing that emergent abilities may be a measurement artifact, we can refine our tools to detect nuances in text generation. This improved discernment is vital for areas such as content verification, academic integrity, and the identification of misinformation online.

\section{Conclusion}

We systematically evaluated three types of solutions (Transformer, ML, and LLM) using open-source datasets from Reddit, examining the feasibility and limitations of each approach with model and pipeline modifications. Additionally, we reviewed the paper \textit{Are Emergent Abilities of Large Language Models a Mirage}~\cite{NEURIPS2023_adc98a26}, which provides a detailed analysis of this research field.

\bibliographystyle{ACM-Reference-Format}
\bibliography{sample-base}

\appendix









\end{document}